# Efficient and Practical Stochastic Subgradient Descent for Nuclear Norm Regularization

**Haim Avron** HAIMAV@US.IBM.COM
**Satyen Kale** SCKALE@US.IBM.COM
**Shiva Prasad Kasiviswanathan** SPKASIVI@US.IBM.COM
**Vikas Sindhwani** VSINDHW@US.IBM.COM
IBM T.J. Watson Research Center, Yorktown Heights, NY 10598 USA

## Abstract

We describe novel subgradient methods for a broad class of matrix optimization problems involving nuclear norm regularization. Unlike existing approaches, our method executes very cheap iterations by combining low-rank stochastic subgradients with efficient incremental SVD updates, made possible by highly optimized and parallelizable dense linear algebra operations on small matrices. Our practical algorithms always maintain a low-rank factorization of iterates that can be conveniently held in memory and efficiently multiplied to generate predictions in matrix completion settings. Empirical comparisons confirm that our approach is highly competitive with several recently proposed state-of-the-art solvers for such problems.

## 1. Introduction

We consider the following convex optimization problem over matrices:

$$\min_{X \in \mathbb{R}^{m \times n}} f(X) + \lambda \|X\|_*, \tag{1}$$

where $f(X)$ is any convex function (not necessarily differentiable), $\lambda > 0$ is a regularization parameter, and $\|X\|_*$ denotes the nuclear (trace) norm of a matrix $X$, which is the sum, or equivalently the $l_1$ norm, of the singular values of $X$. We assume without loss of generality that $m \geq n$. This setup generalizes, from vectors to matrices, the widely successful idea of using $l_1$-regularization as a convex proxy for imposing sparsity constraints. Sparsity in the spectrum of a matrix corresponds to low-rankness, a key modeling idea that is naturally justified in a variety of application contexts, e.g., recommender systems, topic models, and multi-task learning.

Two broad lines of research have emerged around the analysis and implementation of matrix estimation algorithms based on nuclear norm regularization. One concerns the theoretical characterization of the conditions under which an unknown low-rank matrix can be exactly recovered by solving a problem of the form (1), or variations thereof, given a set of partially observed entries with respect to which the function $f$ provides a measure of prediction quality. A complementary line of work, which this paper contributes to, concerns the development of algorithmic frameworks to efficiently solve (1) for large-scale problems. Given the definition of the nuclear norm, the Singular Value Decomposition (SVD) tends to unsurprisingly play a critical computational role in the design of nuclear norm solvers, e.g., the Singular Value Thresholding (SVT) (Cai et al., 2010), Soft-Impute (Mazumder et al., 2010), accelerated Proximal Gradient approach (Ji & Ye, 2009) and related efforts, all involve applying a soft-thresholding operator on the singular values of an iterate, which requires repeated calls to an SVD solver. In particular, if the iterates need to pass through a region where the spectrum is dense, these techniques can potentially become prohibitively expensive. An alternative approach is proposed by Jaggi & Sulovský (2010) who map (1) to the problem of optimizing a convex function over the set of positive semi-definite matrices with unit trace, for which the Sparse-SDP solver of Hazan (2008) is invoked. This approach is appealing since each iteration involves the computation of only the largest singular value, and associated left and right singular vectors, of the gradient of $f$ at the current iterate which involves relatively cheap sparse-matrix operations followed by quick rank-one updates. On the other hand, this ap-proach produces an $\epsilon$-accurate solution with rank possibly as large as $\Theta(\frac{1}{\epsilon})$ making it challenging to hold the factorization of the solution in memory and apply it in practice for generating predictions. Another class of methods, e.g. (Recht & Ré, 2011), works with a low-rank parameterization directly; in general, this leads to a non-convex formulation whose solutions may be highly sensitive to initialization.

The contributions of this paper are as follows:

- A new stochastic subgradient descent approach to solving (1). By utilizing a novel subgradient probing technique that generates low-rank steps, combined with enforcing low-rankness in the iterates, our method is able to do cheap iterates and produce a low-rank solution. Furthermore, instead of using sparse SVD as the main computational kernel, our method uses highly scalable dense matrix operation (QR factorization of a tall-and-skinny matrix) as its basic kernel, so it is better poised to take advantage of the computational power of modern platforms.
- While the theoretical worst-case running time of our basic algorithm is $O\left(mn^2\epsilon^{-2}\right)$, an efficient variant that enforces low-rankness shows, empirically, linear complexity in $m$, $n$ and $r$, where the rank of the solution is at most $r$.
- We apply our method to matrix completion and show that it compares favorably to state-of-the-art techniques for solving (1) (Jaggi & Sulovský, 2010; Mazumder et al., 2010; Shalev-Shwartz et al., 2011). In addition, our approach is more general than these methods in that it applies, without any major modifications, to the broader class of subdifferentiable elementwise loss functions.

### 1.1. Preliminaries

We use $i:j$ to denote the set $\{i,\ldots,j\}$, and $[n]=1:n$. Vectors are always column vectors and are denoted by boldface letters. We use $\mathbf{0}_{m\times n}$ to denote an $m\times n$ matrix of all zeros. Given $\mathcal{R}\subseteq[m]$ and $\mathcal{C}\subseteq[n]$, we denote by $X_{\mathcal{R},\mathcal{C}}$ the submatrix of $X$ consisting of rows in $\mathcal{R}$ and columns in $\mathcal{C}$. For a matrix $X\in\mathbb{R}^{m\times n}$ with $m\geq n$, $\sigma(X)=(\sigma_1(X),\ldots,\sigma_n(X))$ is the vector of singular values of $X$, with entries in non-increasing order. The nuclear (trace) norm $\|X\|_*$ of a matrix $X$ is the sum of the singular values of $X$. For a matrix $X$, let $\operatorname{rank}(X)$ denote its rank. When the referenced matrix $X$ is clear, we just use rank to denote $\operatorname{rank}(X)$.

Given the SVD of a matrix $A\in\mathbb{R}^{m\times n}$ with $m\geq n$, viz. $A=U\Sigma V^\top$, the *reduced* SVD factorization consists of discarding the last $m-n$ columns of $U$ and bottom $m-n$ rows of $\Sigma$. By removing singular triplets associated with zero singular values we get the *compact* SVD. *Truncating* a matrix to rank $t$ (or *truncated* SVD), denoted by $\mathrm{TSVD}(A,t)$ consists of removing the $n-t$ smallest singular values from the reduced SVD. It is well known that $\mathrm{TSVD}(A,t)$ is the best rank $t$ approximation to $A$ (spectral and Frobenius norm).

A QR *factorization* of a matrix $A\in\mathbb{R}^{m\times n}$ is decomposition of $A$ into a product $A=QR$ of an unitary matrix $Q\in\mathbb{R}^{m\times m}$ and an upper triangular matrix $R\in\mathbb{R}^{m\times n}$. If $m\geq n$, then the *reduced* QR factorization consists of discarding the last $m-n$ columns of $Q$ and bottom $m-n$ rows of $R$. The QR factorization can be computed efficiently and in a stable manner using $O(mn^2)$ operations.

We assume the following properties of the function $f$, which are quite basic and easily satisfied in most applications:

1. A subgradient of $f$ at any point $X$, $\nabla f(X)\in\partial_X f(X)$, is efficiently computable.
2. We know an upper bound $\Delta$ on $\|X_{\text{opt}}\|_{\text{F}}$.
3. We know an upper bound $G$ on $\|\nabla f(X)\|_{\text{F}}$ for all $X$ such that $\|X\|_{\text{F}}\leq\Delta$.

## 2. Stochastic Subgradient Descent

We now describe our stochastic subgradient descent (SSGD) algorithm for solving nuclear norm regularized problems of type (1). Let $F(X)=f(X)+\lambda\|X\|_*$. Instead of solving (1), we solve

$$\min_{X\in\mathcal{K}} f(X)+\lambda\|X\|_*, \tag{2}$$

where $\mathcal{K}=\{X\in\mathbb{R}^{m\times n}:\ \|X\|_{\text{F}}\leq\Delta\}$. Note that $X_{\text{opt}}\in\mathcal{K}$, so this is problem is equivalent to (1). The reason we use this set $\mathcal{K}$ is to make sure our iterates are bounded. This is ensured by projecting on $\mathcal{K}$ if we ever step outside of $\mathcal{K}$. The projection onto $\mathcal{K}$ is defined as follows:

**Definition 1** (Projection Operator for $\mathcal{K}$)**.** *Define* $\Pi_{\mathcal{K}}(P)=\operatorname{argmin}_{Q\in\mathcal{K}}\|P-Q\|_{\text{F}}=\min\{1,\frac{\Delta}{\|P\|_{\text{F}}}\}P$.

Let $X=U\Sigma V^\top$ be a compact SVD of $X$. Let $U_{\text{rank}},V_{\text{rank}}$ be $U,V$ truncated to the first $\operatorname{rank}(X)$ columns. It is well known that $U_{\text{rank}}V_{\text{rank}}^\top$ is a subgradient of $\|X\|_*$ (Watson, 1992), so we find that a subgradient of $F(X)=f(X)+\lambda\|X\|_*$ is

$$\mathcal{G}(F(X))\stackrel{\text{def}}{=}\nabla f(X)+\lambda\cdot U_{\text{rank}}V_{\text{rank}}^\top\in\partial_X F(X)\,.$$

This gives the following upper bound on $\|\mathcal{G}(F(X))\|_{\text{F}}$ for any $X\in\mathcal{K}$:

$$\begin{aligned}\|\mathcal{G}(F(X))\|_{\text{F}} &\leq \|\nabla f(X)\|_{\text{F}}+\lambda\|U_{\text{rank}}V_{\text{rank}}^\top\|_{\text{F}}\\ &\leq G+\lambda\sqrt{\text{rank}}\,.\end{aligned}$$

The crucial ingredient for a SSGD algorithm is an unbiased estimator for a subgradient. Our technique for computing such an estimator is to *probe* $\mathcal{G}(F(X))$ by multiplying it by a random matrix.

**Definition 2** (Probing Matrix). *A random* $n \times k$ *matrix* $Y$ *is a* probing matrix *if* $\mathbb{E}[YY^\top] = \mathrm{I}_{n\times n}$ *where* $\mathrm{I}_{n\times n}$ *is the* $n \times n$ *identity matrix and the expectation is over the choice of* $Y$.

Here, $k \ll n$ is a parameter that is adjustable in our algorithm. The following lemma gives several families of distributions that generate probing matrices efficiently. (All missing proofs appear in the full version of the paper (Avron et al., 2012).)

**Lemma 2.1.** *Let* $Y = Z/\sqrt{k}$ *where* $Z$ *is a random matrix drawn any one of the following distributions:*

1. *Independent entries taking values* $+1$ *and* $-1$ *with equal probability* $1/2$.
2. *Independent and identically distributed standard normal entries.*
3. *Each column of* $Z$ *is drawn uniformly at random and independent of each other from* $\{\sqrt{n}e_1, \ldots, \sqrt{n}e_n\}$ *(scaled identity vectors).*

*Then* $Y$ *is a probing matrix.*

By linearity of expectation, for any matrix $A$ we have $\mathbb{E}[AYY^\top] = A\,\mathbb{E}[YY^\top] = A$. Thus, in our algorithms we use $\mathcal{G}(F(X))YY^\top$ as an unbiased estimator of $\mathcal{G}(F(X))$. Now, using an unbiased estimator instead of an exact subgradient (that is, using SSGD instead of subgradient descent) is beneficial only if there is some computational advantage in doing so. Our probing technique has two potential advantages over the use of an exact subgradient. First, for any matrix $A$ the matrix $AYY^\top$ has rank at most $k$. So our estimator is, in fact, a *low-rank* unbiased estimator. We will utilize this low-rankness later on. Second, often it is more efficient to compute the product $\mathcal{G}(F(X))Y$ than to actually compute $\mathcal{G}(F(X))$. For example, if $Y$ is composed of scaled identity vectors (case 3 in Lemma 2.1), then $\mathcal{G}(F(X))Y$ is a matrix composed of $k$ columns of $\mathcal{G}(F(X))$, so we need to compute only a small portion of $\mathcal{G}(F(X))$. Henceforth, we use the scaled identity vectors as the default probing matrix.

### 2.1. Basic SSGD Algorithm

Using the above ingredients, we now describe a basic SSGD algorithm for solving (2). We use $g^{(t)} = \mathcal{G}(F(X^{(t)}))YY^\top$ as an unbiased estimator of $\mathcal{G}(F(X^{(t)}))$. Algorithm Basic-SSGD describes the complete procedure. Basic-SSGD is not efficient in terms of running time and memory requirements. Subsequent subsections give more efficient variants.

**Algorithm 1** Basic-SSGD

**Input:** $f$, $\lambda$, $T$, step sizes $\eta^{(1)}, \ldots, \eta^{(T-1)}$, and $k$

Initialize $X^{(0)} = \mathbf{0}_{m\times n}$
**for** $t = 0$ **to** $T-1$ **do**
  Generate an $n \times k$ probing matrix $Y$
  $g^{(t)} \longleftarrow \mathcal{G}(F(X^{(t)}))YY^\top$
  $X^{(t+1)} \longleftarrow \Pi_{\mathcal{K}}(X^{(t)} - \eta^{(t)} g^{(t)})$
**end for**
Return $X^{(\ell)} = \mathrm{argmin}_{X^{(t)}, 0\le t\le T} F(X^{(t)})$

We now analyze Algorithm Basic-SSGD.

**Lemma 2.2.** *Let* $Y \in \mathbb{R}^{n\times k}$. *Then for any matrix* $A \in \mathbb{R}^{m\times n}$,

$$\mathbb{E}[\|AYY^\top\|_{\mathrm{F}}^2] = \mathrm{trace}(\mathbb{E}[YY^\top YY^\top]A^\top A).$$

*In particular, if* $Y$ *is a probing matrix from Definition 2, then* $\mathbb{E}[\|AYY^\top\|_{\mathrm{F}}^2] = \frac{n}{k}\|A\|_{\mathrm{F}}^2$.

Substituting $A = \mathcal{G}(F(X^{(t)}))$ in Lemma 2.2 gives

$$\begin{aligned}\mathbb{E}[\|g^{(t)}\|_{\mathrm{F}}^2] &\le \frac{n}{k}\|\mathcal{G}(F(X^{(t)}))\|_{\mathrm{F}}^2 \\ &\le \frac{n}{k}\left(G + \lambda\sqrt{\mathrm{rank}(X^{(t)})}\right)^2.\end{aligned}$$

Finally, using this in a standard bound on the convergence rate of the SSGD algorithm , we get the following theorem about the convergence rate of the Basic-SSGD algorithm.

**Theorem 2.3.** *Suppose* $r$ *is an upper bound on* $\mathrm{rank}(X^{(t)})$ *for all* $t$. *Then if we set* $\eta^{(t)} = \beta\frac{\sqrt{k}\Delta}{\sqrt{n}(G+\lambda\sqrt{r})\sqrt{T}}$, *the solution* $X^{(l)}$ *of Algorithm* Basic-SSGD *satisfies*

$$\mathbb{E}[X^{(l)}] - F(X_{\mathrm{opt}}) \le 4\sqrt{n}\frac{(G+\lambda\sqrt{r})\Delta}{\sqrt{kT}}\max\left\{\beta, \frac{1}{\beta}\right\}.$$

Thus Basic-SSGD converges to within $\epsilon \cdot \|X_{\mathrm{opt}}\|_{\mathrm{F}}$ of the optimal goal in $O(n/k\epsilon^2)$ iterations. If we fix $\epsilon$ then convergence is in $O(n/k)$ iterations.

### 2.2. Decreasing the Iteration Cost of Basic-SSGD for Low Rank Iterates

In each iteration of Basic-SSGD, we have to compute $g^{(t)}$. The straightforward way of computing $g^{(t)}$ requires computing the SVD of $X^{(t)}$, which has $O(mn^2)$ time complexity and is computationally prohibitive for large matrices. Fortunately, when the rank of $X^{(t)}$ is small we can do much better. Let $r^{(t)}$ denote the rank of $X^{(t)}$. Assume that we have a compact SVD decomposition of $X^{(t)}$: $X^{(t)} = U^{(t)}\Sigma^{(t)}(V^{(t)})^\top$. Since $X^{(0)} = \mathbf{0}_{m\times n}$, we certainly have such a decomposition for $X^{(0)}$. We now show how to generate the SVD of $X^{(t+1)}$, without explicitly forming $X^{(t+1)}$. To do so, we first define a class of loss functions.

**Definition 3** (Sum Loss Function)**.** *A function $f : \mathbb{R}^{m\times n} \to \mathbb{R}$ is a* sum loss function *if for $X \in \mathbb{R}^{m\times n}$, $f(X) = \sum_{ij} f_{ij}(X_{ij})$ where the $f_{ij}$s are convex functions. We say that $f$ is* efficiently computable *if $f_{ij}(x)$ and a subgradient $f'_{ij}(x)$ of $f_{ij}$ at $x$ can be computed in $O(1)$ time for every $i$, $j$, and $x$.*

Note that the above definition captures many commonly used functions, including squared error, absolute error, hinge loss function (with or without smoothing), and can also apply to arbitrary domain-specific subdifferentiable elementwise loss functions.

**Lemma 2.4.** *Suppose that $f$ is an efficiently computable sum loss function. Suppose that the probing matrix $Y$ is composed of scaled identity vectors. Given the SVD of $X^{(t)}$, there is an algorithm to compute the SVD of $X^{(t+1)} = \Pi_{\mathcal{K}}(X^{(t)} - \eta^{(t)} g^{(t)})$ (without explicitly computing $X^{(t+1)}$ itself) in $O(m(r^{(t)}+k)^2)$ time, where $r^{(t)}$ is the rank of $X^{(t)}$.*

*Proof.* We first write a formula for $g^{(t)}$:

$$\begin{aligned} g^{(t)} &= \mathcal{G}(X^{(t)})YY^\top \\ &= \underbrace{[\nabla f(U^{(t)}\Sigma^{(t)}(V^{(t)})^\top)Y + U^{(t)}(V^{(t)})^\top Y]}_{S^{(t)}} Y^\top . \end{aligned}$$

By first computing $(V^{(t)})^\top Y$ and then multiplying by $U^{(t)}$ we can compute $U^{(t)}(V^{(t)})^\top Y$ in $O(mkr^{(t)})$ time.

Since $Y$ is composed of scaled identity vectors, $\nabla f(X^{(t)})Y$ is composed of scaled columns of $\nabla f(X^{(t)})$. The expansion $f(X) = \sum_{ij} f_{ij}(X_{ij})$ implies that $(\nabla f(X))_{ij} = f'_{ij}(X_{ij})$, so to find $\nabla f(X^{(t)})Y$ we simply have to compute the values $\nabla f(X^{(t)})$ at the corresponding columns. The entries of a single column of $X^{(t)}$ can be computed from the decomposition $X^{(t)} = U^{(t)}\Sigma^{(t)}(V^{(t)}))^\top$ in $O(mr^{(t)})$ time, so $k$ columns can be computed in $O(mkr^{(t)})$. Therefore, $S^{(t)}$ can be computed in $O(mkr^{(t)})$ time.

Since $g^{(t)} = S^{(t)}Y^\top$ we have $X^{(t+1)} = \Pi_{\mathcal{K}}(X^{(t)} - \eta^{(t)}S^{(t)}Y^\top)$. Let us now define

$$\widehat{U}^{(t+1)} \stackrel{\text{def}}{=} [U^{(t)}\Sigma^{(t)} \;\; S^{(t)}] \text{ and } \widehat{V}^{(t+1)} \stackrel{\text{def}}{=} [V^{(t)} \;\; -\eta^{(t)}Y]$$

and we get that $X^{(t+1)} = \Pi_{\mathcal{K}}(\widehat{U}^{(t+1)}(\widehat{V}^{(t+1)})^\top)$. This is already a low-rank representation of the matrix. We now show how to turn the low-rank representation $\widehat{U}^{(t+1)}(\widehat{V}^{(t+1)})^\top$ to a compact SVD. First, we compute a reduced QR decomposition of $\widehat{U}^{(t+1)} = Q_U R_U$ and $\widehat{V}^{(t+1)} = Q_V R_V$. We then compute $R_U R_V^\top$ and find an SVD decomposition $R_U R_V^\top = M\bar{\Sigma}^{(t+1)}N^\top$. Then we compute $\bar{U}^{(t+1)} = Q_U M$ and $\bar{V}^{(t+1)} = Q_V N$ and we have

$$X^{(t+1)} = \Pi_{\mathcal{K}}(\bar{U}^{(t+1)}\bar{\Sigma}^{(t+1)}(\bar{V}^{(t+1)})^\top) .$$

The decomposition of the inner matrix is, in fact, a reduced SVD decomposition since $Q_U$, $M$, $Q_V$ and $N$ are orthonormal, and $\bar{\Sigma}^{(t+1)}$ is diagonal with positive values. Since it is a SVD we can efficiently do the $\mathcal{K}$ projection and remove zero (or numerically zero) singular values to obtain the compact SVD decomposition $X^{(t+1)} = U^{(t+1)}\Sigma^{(t+1)}(V^{(t+1)})$ of the next iterate.

It is easy to verify that the dominant operation, in terms of running time, is the computation of the QR factorization of $\widehat{U}^{(t+1)}$. Since this is a factorization of an $m \times (r^{(t)} + k)$ matrix, the overall cost is $O(m(r^{(t)}+k)^2)$, which is much better than $O(mn^2)$ if $r^{(t)} + k \ll n$. For a pseudo-code description, see full version of the paper (Avron et al., 2012). ☐

Unfortunately, in Basic-SSGD it is quite probable that rank of $X^{(t+1)}$ is $r^{(t)}+k$, and therefore, after $n/k$ iterations the iterates could become full rank, and from there on updates will take $O(mn^2)$ time per iteration (same iteration cost as the trivial implementation of Basic-SSGD). We will see how to avoid this situation in the next subsection.

**Scalability.** The running time of each iteration is dominated by computing QR factorizations of tall-and-skinny dense matrices. We could have used SVD computations instead of QR computations. Since QR and SVD have equivalent asymptotic running this would not have changed the time complexity of the algorithm. However, we chose to use QR factorizations instead since it is a simpler operation that exhibits better running time *in practice*. The use of QR factorization also makes the update operation highly scalable on modern parallel architectures. By using a highly tuned package like ScaLAPACK (Choi et al., 1992) good speedups should be attainable with little effort. Recent research on QR factorization has shown how to implement it efficiently on communication-bound massive parallel machines (Demmel et al., 2008), MapReduce clusters (Constantine & Gleich, 2011), and GPUs (Anderson et al., 2011). It is worth noting that our algorithm avoids the computation of singular values on large sparse matrices, which require more sophisticated communication-avoiding methods (Hoemmen, 2010), which do not scale as well as dense linear algebra operations.

### 2.3. Enforcing Low-Rank Iterates

The update algorithm of Lemma 2.4 could be used in Algorithm BASIC-SSGD to go from the SVD of $X^{(t)}$ to SVD of $X^{(t+1)} = \Pi_{\mathcal{K}}(X^{(t)} - \eta^{(t)} g^{(t)})$. The discussion in the previous section shows that if the iterates ($X^{(t)}$s) in Algorithm BASIC-SSGD are low rank then the iterations are fast. This suggests the idea of explicitly truncating the least singular values of the iterates to ensure that the iterates always remain low rank. In this section, we formalize this idea.

Let $\mathcal{M}_r = \mathcal{M}_r^{m \times n}$ denote the set of $m$-by-$n$ matrices of rank at most $r$. Suppose we assume that $\text{rank}(X_{\text{opt}}) \le r$, i.e., $X_{\text{opt}} \in \mathcal{M}_r$, for some parameter $r$. Then minimizing $F(X)$ over $\mathcal{K} \cap \mathcal{M}_r$ yields the same optimum solution. We look at the problem of solving (2) with the additional constraint that $X \in \mathcal{M}_r$. However, the set $\mathcal{M}_r$ is non-convex. Furthermore, rank constraints, like $X \in \mathcal{M}_r$, typically result in NP-hard problems (see (Natarajan, 1995)).

Our strategy is to again use a projected subgradient method. That is, in each iteration we start with a regular stochastic subgradient step. This step takes us out of $\mathcal{M}_r$. The following step is to project back to $\mathcal{M}_r$. It is well known that for any matrix $X$ the best rank $r$ approximation to $X$ (measured in terms of Frobenius norm) can be computed by truncating the SVD to the top $r$ singular values. Since our algorithm keeps a SVD representation of the iterates $X^{(t)}$, we can compute the projection on $\mathcal{M}_r$ efficiently. Despite not having a theoretical guarantee because of the non convexity of $\mathcal{M}_r$, our experiments, which we report in Section 4, suggest that $O(n/k)$ iterations are still sufficient for this explicit rank-constrained nuclear norm regularized problem. Note that the projection operator on $\mathcal{K}$ does not change the rank of the matrix since it is a simple scaling. For a pseudo-code description, see full version of the paper (Avron et al., 2012).

**Role of $r$.** Since nuclear norm regularization is typically used as a proxy for rank constraints, one might wonder why we impose an explicit rank constraint. Primarily, we use the rank constraint only for computational efficiency reasons though we consistently observed statistical benefits from additionally keeping the nuclear norm regularizer (i.e., $\lambda > 0$). As long as $r \ge \text{rank}(X_{\text{opt}})$ the solution to the problem does not change, and the rank constraint is *passive*. So we need only an upper bound on the rank of the optimal solution. We can even select $r = n$, which will remove the rank constraint completely. However, the running time of the algorithm does depend on $r$, so it is best to set $r$ as close as possible to $\text{rank}(X_{\text{opt}})$. Since we use nuclear norm regularization we expect the rank of $X_{\text{opt}}$ to be small.

**Algorithm 2** SSGD-MATRIX-COMPLETION

**Input:** $Z \in \mathbb{R}^{m \times n}$ and parameters $r$, $s$, $\delta$, and $\nu$
- $[U^{(0)}, \Sigma^{(0)}, V^{(0)}] = \text{TSVD}(Z, r)$
- $\alpha \leftarrow \frac{1}{\|Z\|_{\text{F}}^2}$, $\beta \leftarrow \frac{\delta \cdot f(X^{(0)})}{(\|Z\|_{\text{F}}^2 \cdot \|X^{(0)}\|_*)}$
- $\Delta \leftarrow \alpha\beta^{-1}\|Z\|_{\text{F}}$, $\eta \leftarrow \nu\|Z\|_{\text{F}}^2$
- **for** $t = 0$ **to** $s\lceil \frac{n}{r} \rceil$ **do**
  - Create $\mathcal{C} = c_1, \ldots, c_r$, $\forall i$ $c_i$ drawn i.i.d. from $[n]$
  - $Y \leftarrow [e_{c_1}, \ldots, e_{c_r}]$
  - $P^{(t)} \leftarrow \beta U^{(t)} (V^{(t)}_{\mathcal{C},1:r})^\top$
  - $S^{(t)} \leftarrow \sqrt{\frac{n}{k}}(2\alpha(U^{(t)}\Sigma^{(t)}(V^{(t)}_{\mathcal{C},1:r})^\top - Z_{1:m,\mathcal{C}}) + P^{(t)})$
  - $\widehat{U}^{(t+1)} \leftarrow [U^{(t)}\Sigma^{(t)} \;\; S^{(t)}]$
  - $\widehat{V}^{(t+1)} \leftarrow [V^{(t)} \;\; -\eta^{(t)} Y]$
  - Factorize: $\widehat{U}^{(t+1)} = Q_U R_U$, $\widehat{V}^{(t+1)} = Q_V R_V$
  - $T \leftarrow R_U R_V^\top$
  - SVD computation: $T = M \bar{\Sigma}^{(t+1)} N^\top$
  - $\bar{U}^{(t+1)} \leftarrow Q_U M$, $\bar{V}^{(t+1)} \leftarrow Q_V N$
  - $U^{(t+1)} \leftarrow \bar{U}^{(t+1)}_{1:m,1:r}$, $V^{(t+1)} \leftarrow \bar{V}^{(t+1)}_{1:n,1:r}$
  - $\Sigma^{(t+1)} \leftarrow \bar{\Sigma}^{(t+1)}_{1:r,1:r}$
  - **if** $\|\Sigma^{(t+1)}\|_{\text{F}} > \Delta$ **then**
    - $\Sigma^{(t+1)} \leftarrow \Sigma^{(t+1)}\Delta / \|\Sigma^{(t+1)}\|_{\text{F}}$
  - **end if**
- **end for**

**Output.** The output of our algorithm matrix in compact SVD form. This is a much more succinct representation that requires only $O(mr)$ memory words, instead of $O(mn)$. A specific entry in the matrix can be computed in $O(r)$ time, and the entire matrix can be computed in $O(mnr)$ time.

**Choice of $k$.** Since we are explicitly enforcing the rank of the iterates to be smaller than $r$ we have $r^{(t)} \le r$. This implies that each iteration is computed in $O(m(r+k)^2)$ time. There is a relation between $k$ and the number of iterations: as $k$ grows our gradients improve so we expect to converge in less iterations. In all our experiments we set $k = r$ to avoid an additional tunable parameter

## 3. Application: Matrix Completion

In the low rank matrix completion problem, we are given a set of indices $\Omega \in [m] \times [n]$ and associated values $(Z_{ij})_{(i,j)\in\Omega}$ and we are required to complete the matrix with the lowest possible rank. Minimizing the rank is hard, so a popular approach for solving the matrix completion is as follows. Let $\mathcal{P}_\Omega$ be the projection onto the index set $\Omega$. That is, $(\mathcal{P}_\Omega(X))_{ij} = X_{ij}$ if $(i,j) \in \Omega$ and 0 otherwise. Let $Z$ be the matrix containing the known values at their correct position and 0 in all other positions. The problem to be solved is then

$$\min_{X \in \mathbb{R}^{m \times n}} \alpha \|\mathcal{P}_\Omega(X) - Z\|_{\mathrm{F}}^2 + \beta \|X\|_* \,. \tag{3}$$

For reasons that will be apparent later we chose to write the problem with two parameters ($\alpha$ and $\beta$) instead of a single parameter $\lambda$. Other variants are possible, like using $\ell_1$-loss instead of $\ell_2$-loss, or any subdifferentiable loss function for that matter, but we will focus on (3).

We now show how are approach can be used to solve (3). Set $f(X) = \alpha\|\mathcal{P}_\Omega(X) - Z\|_{\mathrm{F}}^2$. We have $\nabla f(X) = 2\alpha(\mathcal{P}_\Omega(X) - Z)$. Note that $f(X)$ is an efficiently computable sum loss function.

We also need to bound the Frobenius norm of $X_{\mathrm{opt}}$ in order to define the convex set $\mathcal{K}_{\mathrm{F}}$. We observe that

$$\begin{aligned} \|X_{\mathrm{opt}}\|_{\mathrm{F}} \leq \|X_{\mathrm{opt}}\|_* &\leq \beta^{-1} F(X_{\mathrm{opt}}) \\ &\leq \beta^{-1} F(\mathbf{0}_{m\times n}) = \alpha\beta^{-1}\|Z\|_{\mathrm{F}}^2. \end{aligned}$$

Finally, we need to set the step sizes $\eta^{(t)}$. In our experiments, we found that using a fixed step size $\eta$ gave the best results. To make the step size scale free we set $\eta = \nu\|Z\|_{\mathrm{F}}^2$ where $\nu$ is a parameter.

**Heuristics for warm-starting the algorithm and setting the parameters.** Our algorithm can start from any matrix with rank up to $r$ as long as we have a compact SVD of that matrix. We warm-start our algorithm by a rank $r$ truncated SVD of $Z$ (i.e., $X^{(0)} = \mathrm{TSVD}(Z, r)$). The initial SVD is also useful for setting $\alpha$ and $\beta$ in a scale free manner. The nuclear norm serves as a regularizer, so we expect $\beta\|X_{\mathrm{opt}}\|_*$ to be some magnitude smaller than $\alpha f(X_{\mathrm{opt}})$. We do not know the values of $\|X_{\mathrm{opt}}\|_*$ and $f(X_{\mathrm{opt}})$, so instead we use the values of $\|X^{(0)}\|_*$ and $f(X^{(0)})$. First, we set $\alpha = 1/\|Z\|_{\mathrm{F}}^2$, which will make the value of $\alpha f(X)$ range between 0 and 1. We then find $\beta$ such that $\beta\|X^{(0)}\|_* = \delta \cdot \alpha f(X^{(0)})$, where $\delta$ is a new parameter. That is we set $\beta = \delta \cdot f(X^{(0)})/(\|Z\|_{\mathrm{F}}^2 \cdot \|X^{(0)}\|_*)$.

The complete pseudo-code listed in Algorithm SSGD-Matrix-Completion. Overall, Algorithm SSGD-Matrix-Completion has four parameters: (i) bound on the rank of the solution ($r$), (ii) number of super-iterations ($s$; we call every $\lceil \frac{n}{r} \rceil$ iterations a super-iteration), (iii) normalized regularization parameter ($\delta$), and (iv) normalized step size ($\nu$).

## 4. Experimental Results

We ran our matrix completion algorithm (Algorithm SSGD-Matrix-Completion) on two standard collaborative filtering datasets. The first dataset (MovieLens 10M, partition-rb) has about $10^7$ ratings of 69878 users on 10677 movies. The second dataset (Netflix) has about $10^8$ ratings of 480189 users on 17770 movies. The ratings are on an integer scale from 1 to 5. We partitioned each dataset into training and test sets as done by Jaggi & Sulovský (2010). We preprocessed the datasets as follows. For every row and column of training matrix $Z$ we computed the mean. Let $\mu_i$ denote the mean rating of row $i$, and $\hat{\mu}_j$ denote the mean rating of column $j$. We subtract from each training and test rating, $X_{ij}$, the value $(\mu_i + \hat{\mu_j})/2$. In the graphs we refer to our algorithm as "SSGD".

**JSH and Soft-Impute** We compared our results to two other matrix completion algorithms which are also based on solving nuclear norm regularized problems. The first algorithm (which we refer to as "JSH"), suggested by the works of Hazan (2008) and Jaggi & Sulovský (2010), is based on an extension of the Frank-Wolfe (Frank & Wolfe, 1956) algorithm for optimizing a function over the bounded positive semidefinite cone (see also (Clarkson, 2008)). We used a simple MATLAB implementation of the Algorithm 2 from (Jaggi & Sulovský, 2010). The function ApproxEV was implemented by calling MATLAB's **svds** function with default parameters (fixed tolerance). The regularization parameter ($t$) was set according to the best values reported (Jaggi & Sulovský, 2010), i.e., $t = 48333$ for MovieLens and $t = 99592$ for Netflix. The second algorithm that we compared to is the soft singular value thresholding algorithm of Mazumder et al. (2010). We refer to this algorithm as "Soft-Impute". Here, we used the MATLAB code provided by the authors (see (Mazumder et al., 2010)). We used the path-following strategy suggested by the code, i.e., we set a path of values for the regularization parameter ($\lambda$) where the results on a value are used as a warm-start for the next value. The results where examined (RMSE measured) at the different points along the regularization path, and the running time at any point $\lambda_i$ is the sum of time needed to run the algorithm at $\lambda_i$ and the running time of $\lambda_{i-1}$ (because of the warm start). The path used for MovieLens was $(\lambda_0/2, \lambda_0/4, \lambda_0/8, \lambda_0/10)$, and the path used for Netflix was $(\lambda_0/250, \lambda_0/300)$, where $\lambda_0$ is the spectral norm of the input matrix $Z$. Note that both Jaggi & Sulovský (2010) and Mazumder et al. (2010) apply additional heuristics and/or post-processing to their basic algorithm. Additionally, Mazumder et al. (2010) measure only time spent on SVD computations.

These differences in the experimental setup, together with our effort to bring all algorithms under exactly the same experimental protocol, might explain the discrepancy between the results we show here and the results reported in (Jaggi & Sulovský, 2010) and (Mazumder et al., 2010). Nonetheless, for completeness, we also report their published RMSEs. When all algorithms are compared in the same setting, and even relative to best reported RMSEs, we find that our SSGD approach compares favorably to the Frank-Wolfe and singular value thresholding based approaches for solving matrix completion problems.

**Experimental Setup.** We used a 64-bit version of MATLAB 7.8. The experiments were done on a two quad-core Intel E5410 computer running at 2.33 GHz, with 32GB DDR2 800 MHz RAM, running Linux 2.6. None of the codes explicitly uses the eight cores, although some operations (like dense QR factorization) are automatically parallelized by MATLAB. The measured running times are wall-clock times and were measured using the **ftime** Linux system call.

**Setting the parameters and sensitivity to them.** In Figure 1, we examine SSGD-Matrix-Completion's sensitivity to the value of the parameters $r, \delta$, and $\nu$. The best choice of rank ($r$) is 11. However, increasing the rank from 11 to 15 only results in a 0.75% increase in RMSE, and decreasing the rank to 7 only causes a 0.38% increase in RMSE. The best choice of $\delta$ is around 0.01. Overestimating $\delta$ affects the RMSE more adversely than underestimating. For example, setting $\delta$ to 0.1 results in a 5% increase in the RMSE, whereas setting $\delta$ to 0.001 only leads to a 0.48% increase. The best choice of step size ($\nu$) is around 0.009, and the RMSE increases quite smoothly as we go away from this value. We set $\delta = 0.015$, and $\nu = 0.005$ based on preliminary observations on MovieLens 10M without attempting any exhaustive tuning. For Netflix, we simply used the same parameters without any additional experimentation.

**Results on the MovieLens 10M Dataset.** In Figure 2(a), we plot the RMSE on the test set as a function of time. SSGD-Matrix-Completion decreases the error much faster than the other two algorithms, and maintains a better error throughout. SSGD-Matrix-Completion achieved a RMSE of 0.8721 after 1 hour. After running for 180 super-iterations, which took 11.47 hours, the RMSE of SSGD-Matrix-Completion was 0.8555. Compared to this, the JSH obtained a RMSE 0.8640 after 11.66 hours and Soft-Impute obtained a RMSE of 0.8605 after 12.19 hours. Jaggi & Sulovský (2010) report 0.8573 as the best RMSE obtained using their implementation.

In Figure 2(b), we plot the RMSE as a function of rank of the iterates. Every iteration of JSH involves addition of a rank-one matrix, so the rank of iterates soon becomes large. For both JSH and Soft-Impute, we needed to go to a much larger rank to obtain a RMSE comparable to that of a rank-11 solution obtained by SSGD-Matrix-Completion. We stress that at a comparable RMSE a low rank solution is more useful than an high rank solution since it can be held in memory and can be queried much faster to produce a prediction.

**Results on the Netflix Dataset.** Here too, SSGD-Matrix-Completion outperforms JSH and Soft-Impute. After running for 25 super-iterations, which took 23.97 hours, SSGD-Matrix-Completion obtained a RMSE of 0.9516. After 24.08 hours, JSH obtained a RMSE of 0.9583 while Soft-Impute achieved its best RMSE of 0.9603 after 8.55 hours (after 24 hours Soft-Impute's RMSE was worse than this number). After 180 super-iterations, SSGD-Matrix-Completion obtained a RMSE of 0.9411. Jaggi & Sulovský (2010) report 0.9478 as the best RMSE obtained using their optimized implementation. Mazumder et al. (2010) report 0.9497 as the best RMSE obtained in their experiments.

**Comparison with GECO**. The GECO algorithm proposed in Shalev-Shwartz et al. (2011) is not a solver for nuclear norm regularized problems, but uses a greedy method, with optimality guarantees, to optimize $f$ under explicit low-rank constraints. We used the implementation provided to us by the authors with recommended parameters. On MovieLens 10M, GECO returned the best RMSE of 0.8771 at a rank of 17. Our approach yields a better RMSE at a lower rank though GECO's results reaffirm the effectiveness of maintaining explicit low-rank constraints. On the other hand, the running time of the GECO implementation was found to be significantly worse than other methods and we considered it impractical to run it on the Netflix dataset.

**Comparison to Non-Convex Methods.** Methods that work directly with a low-rank parameterization are significantly different in flavor from convex methods since the local minima they find may or may not be optimal for a given choice of rank parameter. In theory, they may be sensitive to initialization and require restarts as reported by Recht & Ré (2011). In practice, on the two datasets that we tested after our draft was written, we found them to be quite robust and efficient, e.g., on MovieLens 10M they get to an RMSE of 0.857 in 45 mins and on Netflix they attain RMSE of 0.9399 in a couple of hours. This is not surprising since these methods, and their variations, dominated the Netflix contest. We advocate that efforts to improve the performance of convex nuclear norm meth-ods, with an eye towards making them efficient and practical, should include such comparisons.

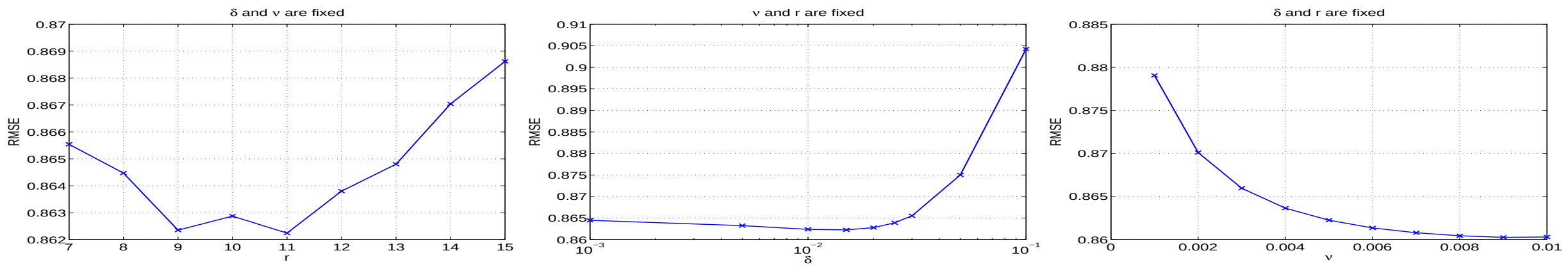

*Figure 1.* Sensitivity of SSGD to parameters on the MovieLens 10M dataset. We run SSGD for 45 super-iterations. In each of the three graphs, we fix two parameters and vary the third.

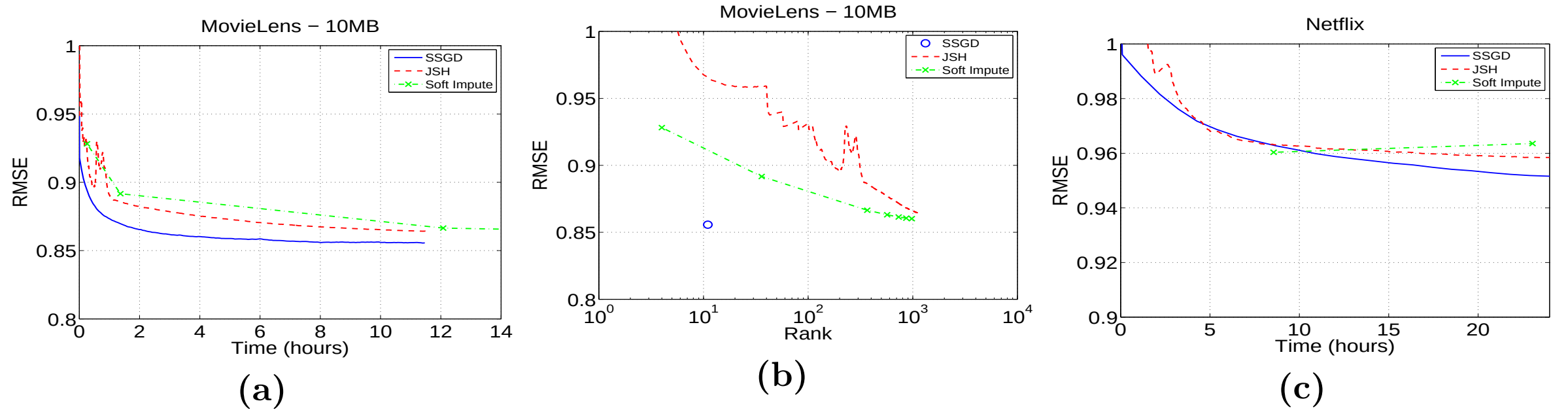

*Figure 2.* Figure (a): test RMSE vs time on MovieLens 10M. Figure (b): test RMSE vs rank MovieLens 10M (for SSGD-MATRIX-COMPLETION, rank $r = 11$; figure 1 shows its RMSE vs rank). Figure (c): shows test RMSE vs time on the Netflix dataset.